\documentclass{article}
\usepackage{amsmath,graphicx,mlspconf}
\usepackage[acronym]{glossaries}
\usepackage{booktabs}
\usepackage{siunitx}
\usepackage{textcomp}
\usepackage{hyperref}
\usepackage{cleveref}
\usepackage{float}

\newacronym{ml}{ML}{Machine Learning}
\newacronym{nas}{NAS}{Neural Architecture Search}
\newacronym{darts}{DARTS}{Differentiable Architecture Search}
\newacronym{pc-darts}{PC-DARTS}{Partial Channel Connections for Memory-Efficient Differentiable Architecture Search}
\newacronym{llm}{LLM}{Large Language Model}
\newacronym{nn}{NN}{Neural Network}
\newacronym{mfcc}{MFCC}{Mel-frequency Cepstral Coefficient}
\newacronym{gsc}{GSC}{Google Speech Commands}

\copyrightnotice{979-8-3503-2411-2/25/\$31.00 {\copyright}2025 European Union}

\toappear{2025 IEEE International Workshop on Machine Learning for Signal Processing, Aug.\ 31-- Sep.\ 3, 2025, Istanbul, Turkey}

\title{Data Aware Differentiable Neural Architecture Search for Tiny Keyword Spotting Applications}
\name{%
    Yujia Shi$^{\star}$%
    \qquad Emil Njor$^{\star}$\thanks{This work is supported by the Innovation Fund Denmark for the project DIREC (9142-00001B).}%
    \qquad Pablo Martínez-Nuevo$^{\dagger}$%
    \qquad Sven Ewan Shepstone$^{\dagger}$%
    \qquad Xenofon Fafoutis$^{\star}$%
}
\address{%
    $^{\star}$ Technical University of Denmark (DTU) \\%
    $^{\dagger}$ Bang \& Olufsen A/S%
}
\begin{document}

\maketitle

\begin{abstract}
The success of Machine Learning is increasingly tempered by its significant resource footprint, driving interest in efficient paradigms like TinyML.
However, the inherent complexity of designing TinyML systems hampers their broad adoption.
To reduce this complexity, we introduce ``Data Aware Differentiable Neural Architecture Search''.
Unlike conventional Differentiable Neural Architecture Search, our approach expands the search space to include data configuration parameters alongside architectural choices.
This enables Data Aware Differentiable Neural Architecture Search to co-optimize model architecture and input data characteristics, effectively balancing resource usage and system performance for TinyML applications.
Initial results on keyword spotting demonstrate that this novel approach to TinyML system design can generate lean but highly accurate systems.
\end{abstract}

\newcommand{\cem}[1]{\textcolor{blue}{cem: #1}}
\section{Introduction}
\label{sec:intro}
Today, \gls{ml} systems are transforming the world, creating exciting applications by turning vast datasets into intelligent systems.
Notable examples include the recent explosion of \glspl{llm}-powered chatbots that can act as personal assistants~\cite{team2023gemini}.
Although these advances are admirable, they happen at the expense of escalating resource consumption of large \gls{ml} systems and ever-higher implications on global environmental challenges.

As \gls{ml} systems permeate more areas of society, resource efficiency becomes essential for the sustainable use of the technology. 
Resource-efficient approaches to \gls{ml} include technologies such as TinyML, where \gls{ml} systems are deployed on low-power systems, promising low energy consumption, network infrastructure independence, and privacy.

Despite its potential, the widespread adoption of TinyML is currently hampered by the technical complexity involved in creating these systems.
Successfully developing TinyML systems demands a rare combination of expertise in designing performant \gls{ml} models, optimizing these models for minimal resource consumption, and implementing them within the constraints of embedded systems.
Because few engineers possess proficiency across all these areas, the development pipeline for novel TinyML applications is often bottlenecked.

AutoML techniques, such as \gls{nas}, which automates the design of \gls{nn} architectures, offer a promising path towards lowering the entrance barrier for creating TinyML systems.
State-of-the-art approaches to \gls{nas} include \gls{darts}~\cite{liu2018darts}, which turn the usually discrete model architecture search space into a continuous one.
This enables the use of gradient-based optimization techniques that can vastly increase the efficiency of the \gls{nas}.
Complementary to this, recent work in TinyML has proposed ``Data Aware \gls{nas}'' which co-optimizes model architecture and input data characteristics to balance resource usage and system performance in TinyML applications~\cite{njor2022primer}.

This work presents early results combining these two \gls{nas} techniques for keyword-spotting applications.
By integrating these approaches, we aim to develop a novel system that capitalizes on the strengths of both: the search efficiency inherent to \gls{darts} and the capacity of Data Aware \gls{nas} to generate lean yet highly accurate TinyML systems\footnote{A code implementation of this system can be found at: \url{https://github.com/ss1k/Data-Aware-NAS}}.

\section{Related Work}
The field of automating the process of creating \gls{ml} systems is commonly known as AutoML~\cite{he2021automl}.
AutoML encompasses various techniques, such as hyperparameter optimization~\cite{yang2020hyperparameter}, automated feature engineering~\cite{nargesian2017learning}, and \gls{nas}~\cite{zoph2017neural}.

\Gls{nas}, the focus of this work, initially relied on discrete black-box optimization techniques such as evolutionary algorithms~\cite{real2017large} and reinforcement learning~\cite{zoph2017neural} to discover well-performing \gls{nn} architectures.
These techniques typically operate by sampling individual architectures, evaluating their performance, and using these evaluations to guide subsequent sampling toward promising regions of the search space.

A significant challenge with these early methods was their computational cost, as evaluating each candidate architecture often required training it from scratch.
This spurred the development of supernet-based \glspl{nas}.
In this paradigm, a large, pre-trained ``supernet'' encapsulates the search space, allowing diverse sub-network architectures to be efficiently extracted and evaluated, often requiring minimal fine-tuning. 
While supernets substantially accelerate architecture evaluation, common supernet-based \glspl{nas} still employ discrete black-box optimization techniques~\cite{brock2018smash} for search space exploration. 
A key limitation of these black-box methods is their inability to understand the landscape of the search space they are exploring, leading to an unguided search --- often relying heavily on iterative sampling around the best-found architectures.
Recently, the idea of \acrfull{darts}~\cite{liu2018darts} has gained popularity, in which the discrete architecture search space is relaxed into a continuous one.
This innovation makes applying efficient gradient-based optimization techniques to the search possible.
Consequently, the search can proceed more strategically, leveraging gradient information to navigate toward optimal architectures, thereby significantly improving efficiency.

In parallel, research focused on TinyML, \gls{nas} techniques have been adapted to generate \gls{nn} architectures that are not only performant, but also resource-efficient~\cite{lin2020mcunet,banbury2021micronets,garavagno2024colabnas}.
This often involves multi-objective optimization, where the search aims to maximize accuracy while minimizing resource consumption indicators such as the number of model parameters.
This extension of \gls{nas} is commonly known as Hardware Aware \gls{nas}~\cite{benmeziane2021comprehensive}.
Hardware Aware \glspl{nas} have served as the basis of several well-known TinyML \gls{nn} architectures, including MCUNet~\cite{lin2020mcunet} and Micronets~\cite{banbury2021micronets}.

A recent evolution within \gls{nas} for TinyML extends beyond architectural and hardware considerations to include data pre-processing configurations in the search space.
This approach, known as Data-Aware \gls{nas}~\cite{njor2023data, njor2025fast}, allows the \gls{nas} optimization process to explicitly manage the allocation of limited resources (e.g., memory, compute) by exploring trade-offs between input data fidelity (e.g., higher resolution, complex features) and model complexity (e.g., depth, width). 
While Data-Aware \gls{nas} has shown promising results in generating efficient specialized models, its validation on standard benchmark datasets is still emerging.

\section{Data Aware Differentiable NAS}\label{sec:data_aware_darts}
This work integrates Data Aware \gls{nas} principles with those of \gls{darts}.
The resulting approach, termed \emph{``Data Aware Differentiable \gls{nas}''}, leverages the gradient-based optimization characteristics of \gls{darts} while incorporating data configurations into the expanded search space.

Our work builds upon the setup described in the original \gls{darts} work~\cite{liu2018darts}, and its extension \gls{pc-darts}~\cite{xu2020pcdartspartialchannelconnections}.
Specifically, we adopt its cell-based model architecture search space and the use of continuous relaxation parameters ($\alpha$ and $\beta$) for gradient-based optimization.
We then extend this framework by incorporating data configurations into the search process.

In the \gls{darts} framework, the model architecture search focuses on identifying the internal structure of two fundamental building blocks: a \emph{normal cell}, which maintains feature map resolution, and a \emph{reduction cell}, which halves the resolution while doubling the feature channels.
The final network is constructed by stacking multiple normal cells, with reduction cells inserted at one- and two-thirds of the network depth.

The optimization process in \Gls{darts} determines each cell's internal wiring and operations. 
To make the discrete search space amenable to gradient descent, DARTS employs a continuous relaxation. 
It does so by assigning a learnable architectural parameter, $\alpha$, to each candidate operation (e.g., convolution, pooling) on every edge within the cell.
Similarly, a $\beta$ parameter weighs the importance of different edges in a cell. 
The magnitude of these parameters indicates the importance of each operation. 
These parameters are optimized alongside the network weights, effectively guiding the search towards promising architectures.

Building upon the \gls{pc-darts} foundation, we introduce the key component of our extension: the integration of a data search space.
This extension allows Data Aware Differentiable \gls{nas} to optimize not only the model architecture but also crucial data pre-processing and representation choices.
The specific options within this space are modality-dependent; for instance, image inputs might involve searching over resolution and color encoding, while audio inputs could involve optimizing sample rates or selecting feature extraction techniques such as \glspl{mfcc}.

\subsection{Data Gamma Parameters}\label{subsec:data_gamma_parameters}
As described in \cref{sec:data_aware_darts}, \gls{darts} relies on continuous relaxation parameters to guide the search toward lower loss.
To expand the ideas of \gls{darts} to the data search space, we introduce a continuous relaxation parameter $\gamma_{d}$ for each possible data configuration $d$.

During training, all input data configurations are combined into a single input weighted by these $\gamma$ parameters.
Namely, all $\gamma$ parameters are put through a \texttt{softmax} function that transforms their raw values into the percentage that the data configuration they represent should contribute towards the combined input sample.

In this way, each $\gamma_{d}$ tracks the importance of data configuration $d$ for the final loss of the \gls{ml} system.
Since our search is guided toward a low loss, a data configuration $d$ that contributes highly to a low loss and a higher model accuracy will see its $\gamma_{d}$ parameter increase and thus will make up a greater part of future combined input samples.

At the end of the \gls{darts} search, the data configuration with the highest $\gamma_{d}$ parameter is chosen as the final data configuration for the \gls{ml} system --- just as the model architecture is chosen in the original \gls{darts}.

We include an early stopping mechanism to increase the efficiency of the data configuration search.
This mechanism stops the data configuration search once the maximum $\gamma$ of all data configurations is double that of the second-highest $\gamma$.

\subsection{Training Details} \label{subsec:training_details}
During the Data Aware Differentiable \gls{nas} search phase, we first perform a warm-up (similar to \gls{darts}) where only model weights $w$ are updated. 
Hereafter, in the main search phase, we alternate between updating $w$ on training data and updating the architecture parameters $\alpha$ and $\beta$ and data gamma parameters $\gamma$ on validation data. 
This alternating update strategy ensures that the joint search over data transformations and architectural configurations converges towards the settings that minimize the overall loss.

During the main search phase, we utilize equation \cref{eq:update_gamma,eq:update_no_gamma} respectively before and after our early stopping mechanism to update our $\alpha$, $\beta$, and $\gamma$ parameters. 

\begin{equation}
    [\alpha_{t+1}, \beta_{t+1}, \gamma_{t+1}] = [\alpha_t, \beta_t, \gamma_t] - \eta_{arch} \nabla_{\alpha,\beta,\gamma} \mathcal{L}_{arch},
\label{eq:update_gamma}
\end{equation}

\begin{equation}
    [\alpha_{t+1}, \beta_{t+1}] = [\alpha_t, \beta_t] - \eta_{arch} \nabla_{\alpha,\beta} \mathcal{L}_{arch}.
\label{eq:update_no_gamma}
\end{equation}
In these equations, $\eta_{arch}$ denotes the learning rate for the architecture parameters, while $\nabla_{\alpha,\beta,\gamma} \mathcal{L}_{arch}$ and $\nabla_{\alpha,\beta} \mathcal{L}_{arch}$ represent the gradients of the architecture loss $\mathcal{L}_{arch}$ with respect to the respective parameters. 

\subsection{Aligning Data Dimensions}
Different data configuration options yield data samples of differing dimensions.
This is a challenge, as a fundamental requirement of \gls{darts} is that input data are aligned to a single dimensionality such that they can be combined into a single input sample as described in \cref{subsec:data_gamma_parameters}.

\begin{figure}
    \centering
    \includegraphics[width=1\linewidth]{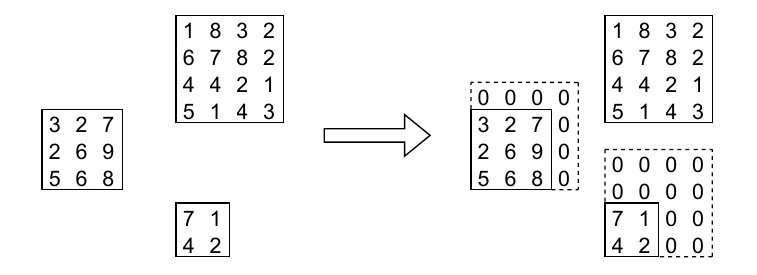}
    \caption{The zero padding strategy to align data dimensionality}
    \label{fig:zero_padding_strategy}
\end{figure}

We propose two strategies for aligning data dimensions: a zero-padding strategy and a pre-processing strategy.
The zero-padding strategy involves zero-padding any lower dimensionality data configuration to the highest dimensionality of all data configurations, as shown in \cref{fig:zero_padding_strategy}.
This strategy preserves all information in all data samples but can require a large amount of processing.
Suppose a lower dimensionality data configuration is found to work the best for a given application. 
In that case, the final \gls{ml} system can chip away all zero padding from both the input dimensions and internal tensors.

\begin{figure}
    \centering
    \includegraphics[width=1\linewidth]{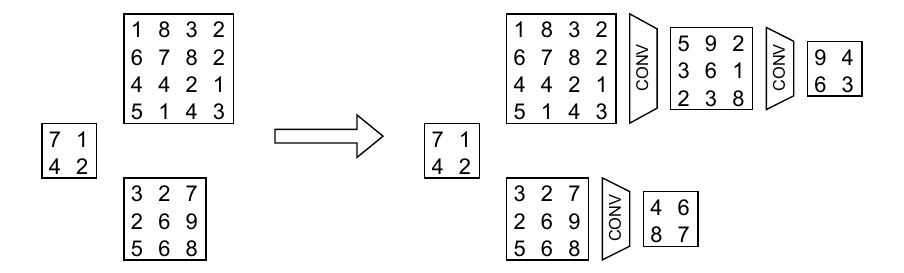}
    \caption{The pre-processing strategy to align data dimensionality}
    \label{fig:pre-processing_strategy}
\end{figure}

The pre-processing strategy involves early processing of higher dimensionality data configurations to align them to the lowest dimensionality of all data configurations, e.g., using convolutional layers as shown in \cref{fig:pre-processing_strategy}.
This strategy can lower processing requirements but may preemptively process important information stored in higher dimensionality data configurations, leading to lower predictive performance.

\section{Results}
We conduct two complementary experiments to demonstrate the utility of Data Aware Differentiable \gls{nas}.
First, we evaluate its overall performance on the \gls{gsc} v0.02 dataset, a standard audio benchmark for TinyML applications~\cite{banbury2021mlperf}, allowing comparison against state-of-the-art systems.
Secondly, we apply Data Aware Differentiable \gls{nas} to a custom name detection task derived from the same dataset, showcasing its flexibility in generating specialized TinyML systems with minimal reconfiguration.
Each experiment follows a two-step process:
\begin{description}
    \item[Search Phase:] Data Aware Differentiable \gls{nas} explores the combined data configuration and neural architecture search space to identify a promising candidate TinyML system configuration.
    \item[Evaluation Phase:] The discovered system configuration (data pre-processing and model architecture) is trained from scratch to determine its final performance metrics.
\end{description}

We experimentally find that the pre-processing alignment strategy achieves better predictive performance on the \gls{gsc} v0.02 dataset.
Therefore, the main results presented here are achieved using the pre-processing alignment strategy.

\subsection{Google Speech Commands}\label{subsec:google_speech_commands_results}
The \gls{gsc} dataset version 0.02 ~\cite{warden2018speech} consists of 105,829 audio files created by 2,618 speakers.
Each file contains the uttering of one of 35 different words lasting at most one second, stored at a sample rate of \SI{16}{\kilo\hertz}.
To align the length of all data samples, we zero-pad each sample shorter than one second.

As this dataset comprises speech audio samples, our data search space will span hyperparameters involved in creating \glspl{mfcc}, reflecting typical engineering choices.
These hyperparameters and their options are described in detail in~\Cref{tab:mfcc_params}.

\begin{figure}[h]
    \centering
    \includegraphics[width=1\linewidth]{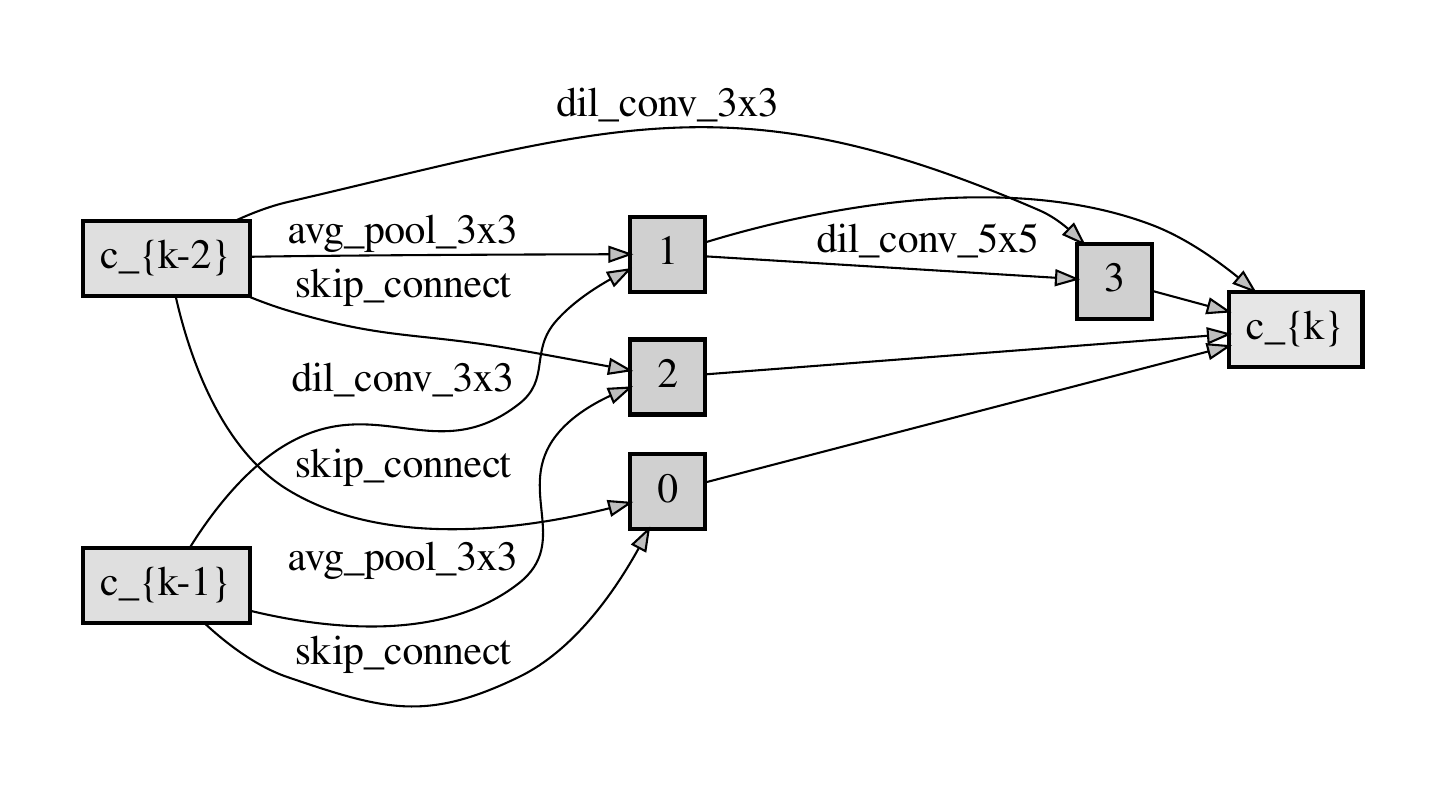}
    \caption{Top-performing normal cell architecture for the \gls{gsc} v0.02 Dataset}
    \label{fig:gsc_conv_normal_cell}
\end{figure}

\begin{figure}
    \centering
    \includegraphics[width=1\linewidth]{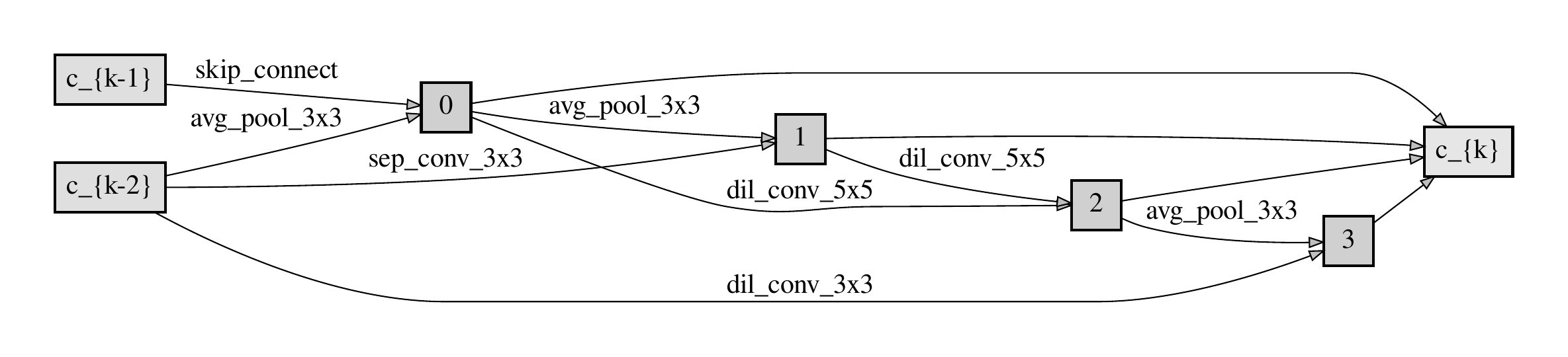}
    \caption{Top-performing reduction cell architecture for the \gls{gsc} v0.02 Dataset}
    \label{fig:gsc_conv_reduction_cell}
\end{figure}

\begin{table}
    \centering
    \caption{MFCC Parameter Search Space}
    \label{tab:mfcc_params}
    \begin{tabular}{lcc}
        \toprule
        \textbf{Window Size} & \textbf{Hop Length} & \textbf{Mel Filters} \\ 
        \midrule
        \textbf{400} & 100 & 40 \\ 
                    & 100 & 80 \\ 
                    & 200 & 40 \\ 
                    & 200 & 80 \\ 
        \midrule
        \textbf{640} & 160 & 40 \\
                    & 160 & 80 \\
                    & 320 & 40 \\ 
                    & 320 & 80 \\ 
        \bottomrule
    \end{tabular}
\end{table}
We run the search phase of Data Aware Differentiable \gls{nas} for 50 epochs on the \gls{gsc} v0.02 dataset, utilizing an 80 GB Nvidia A100 GPU. 
The search used a maximum of 14.59 GB of memory and took approximately 16 hours to complete.
The resulting optimal data configuration has a window size of 400, a hop length of 200, and 40 mel filters.
The discovered normal and reduction cell architectures are illustrated in~ \cref{fig:gsc_conv_normal_cell,fig:gsc_conv_reduction_cell}, respectively
These cells are stacked to yield the final model architecture.


The performance of the discovered system is evaluated by training for 100 epochs, where each epoch iterates once through the full dataset.
\Cref{tab:gsc_evaluation_result} presents our resulting model's accuracy and parameter count alongside several established efficient architectures evaluated using the discovered data configuration and a search that excludes the data-aware component using a data configuration of a window size of 640, a hop length of 160, and 40 mel filters.

\begin{table}
    \centering
    \caption{Parameter count and accuracy of TinyML models on the \gls{gsc} v0.02 dataset configured as described in \cref{subsec:google_speech_commands_results}}
    \label{tab:gsc_evaluation_result}
    \begin{tabular}{lcc}
        \toprule
        \textbf{Model} & \textbf{Parameters} & \textbf{Accuracy}\\
        \midrule
        \textbf{Ours} & \textbf{298 K} & \textbf{97.61\%}\\
        \textbf{Ours} Not Data Aware & 391 K & 96.48\% \\
        DS-CNN~\cite{zhang2017hello} & 1.40 M & 94.68\% \\
        MobileNetV3~\cite{howard2019searching} & 946 K & 82.76\%  \\
        MobileNetV2~\cite{sandler2018mobilenetv2} & 2.27 M & 90.84\% \\
        GhostNet~\cite{han2020ghostnet} & 5.6 M & 92.94\%\\
        EfficientNet~\cite{tan2019efficientnet} & 5.93 M & 88.32\% \\
        \bottomrule
    \end{tabular}
\end{table}

As evident from \cref{tab:gsc_evaluation_result}, our Data Aware Differentiable \gls{nas} achieves higher accuracy than baseline models while utilizing significantly fewer parameters.
This highlights the effectiveness of co-optimizing data pre-processing and architecture for this benchmark.
Note that direct comparison with some highly optimized models reported in the literature (e.g., BC-ResNet~\cite{kim2021broadcasted}) is challenging. 
Such models often achieve state-of-the-art results using distinct training procedures, extensive data augmentation (not employed here), or specialized network blocks potentially unavailable in standard deployment frameworks, complicating fair head-to-head comparisons based solely on reported metrics.

\subsection{Name Detection Application}\label{subsec:name_detection_results}
To demonstrate the flexibility of Data Aware Differentiable \gls{nas}, we adapt the search to work for a different application with minimal framework adjustments.
The application of choice is a name detection application, in which a \gls{ml} model is tasked with detecting utterances of names of people.
Such personalized keyword spotting is relevant for applications like smart headsets that react to a user's name.

We construct a name detection dataset from the \gls{gsc} v0.02 dataset by keeping only the two name classes in the dataset --- ``Marvin'' and ``Sheila'' --- while grouping all other classes into a group of ``unknown'' utterances.
To mitigate the inherent class imbalance, we apply data augmentation techniques, including pitch shifting, time stretching, reverberation, and background noise injection, to achieve a balanced dataset. 
Further details are available in the accompanying code, see the footnote in~\cref{sec:intro}.

\begin{figure}
    \centering
    \includegraphics[width=1\linewidth]{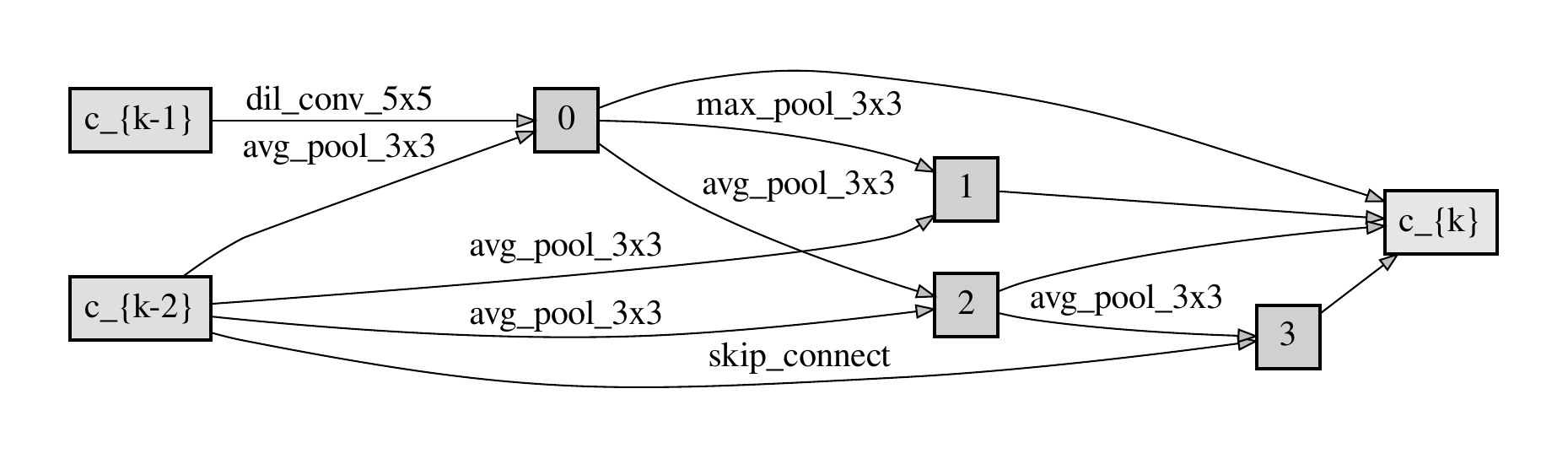}
    \caption{Top-performing normal cell architecture for the name detection application}
    \label{fig:application_conv_normal_cell}
\end{figure}

\begin{figure}[t]
    \centering
    \includegraphics[width=1\linewidth]{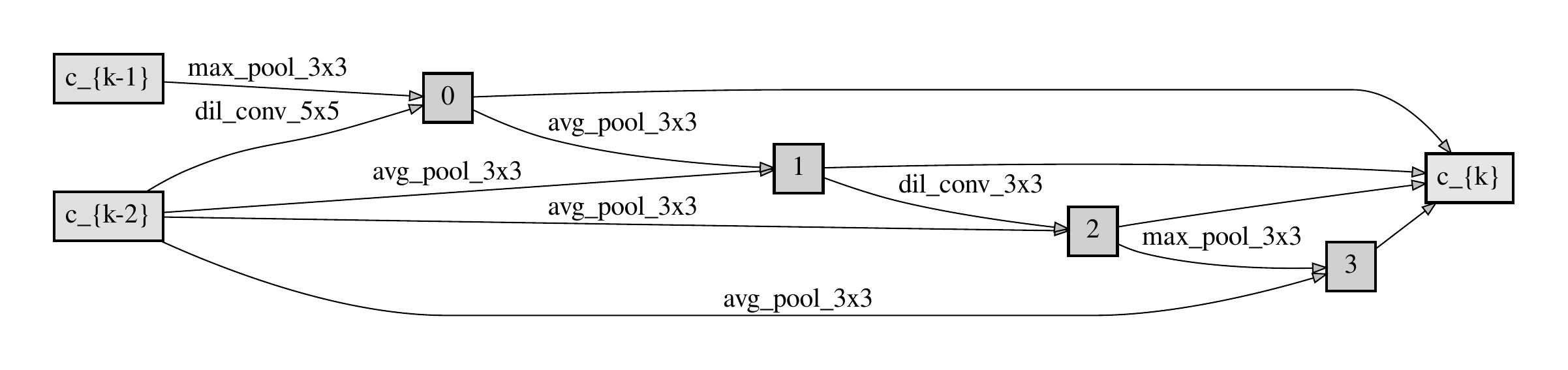}
    \caption{Top-performing reduction cell architecture for the name detection application}
    \label{fig:application_conv_reduction_cell}
\end{figure}

We run the Data Aware Differentiable \gls{nas} search phase for 100 epochs for this name detection application.
This search yields a data configuration with a window size of 640, hop length of 320, and 40 mel filters.
The neural architecture cells found are shown in \cref{fig:application_conv_normal_cell,fig:application_conv_reduction_cell}.


We evaluate the resulting system using the same 100-epoch training procedure applied to the \gls{gsc} v0.02 benchmark. 
\Cref{tab:name_detection_evaluation_result} compares its performance against baseline models using the discovered data configuration. An ablation study that excluded the data search space from the Neural Architecture Search shows a significant drop in performance, it uses a data configuration of a window size of 512, a hop length of 160, and 40 mel filters.

\begin{table}[ht]
\centering
\caption{Parameter count and accuracy of TinyML models for the name detection application with the data configured as described in \cref{subsec:name_detection_results}}
\label{tab:name_detection_evaluation_result}
    \begin{tabular}{lcc}
    \toprule
    \textbf{System} & \textbf{Parameters} & \textbf{Accuracy}\\
    \midrule
    \textbf{Ours} & 1.64M & \textbf{95.43\%}  \\
    \textbf{Ours} Not Data Aware & 2.76M & 91.62\% \\
    DS-CNN~\cite{zhang2017hello} & 1.39M & 93.20\%  \\
    MobileNetV3~\cite{howard2019searching} & \textbf{0.94M} & 84.94\%  \\
    MobileNetV2~\cite{sandler2018mobilenetv2} & 2.26M & 93.03\%  \\
    GhostNet~\cite{han2020ghostnet} & 5.59M & 91.21\%  \\
    EfficientNet~\cite{tan2019efficientnet} & 5.92M & 90.86\%  \\
    \bottomrule
    \end{tabular}
\end{table}

As evident from~\cref{tab:name_detection_evaluation_result}, the model discovered by Data Aware DARTS achieves the highest accuracy (95.43\%) among the evaluated systems.
While MobileNetV3 offers a lower parameter count (0.94M vs. 1.64M), our model significantly outperforms its accuracy.
Our system demonstrates superior accuracy compared to models with similar parameter counts (DS-CNN, MobileNetV2). 
This indicates that while not necessarily the sole Pareto-optimal solution in terms of parameters vs. accuracy across all baselines, Data Aware DARTS successfully identified a configuration yielding top-tier accuracy for this custom task.

\section{Conclusions}
In this paper, we introduced Data Aware Differentiable \gls{nas} --- a novel approach combining the efficiency of \gls{darts} with Data Aware optimization to create resource-efficient but highly accurate TinyML systems.
We extended the foundational PC-DARTS framework to enable the co-optimization of data configurations alongside neural architectures.
Key technical contributions include the introduction of continuous proxy variables to represent data configuration choices within the differentiable search and methods to handle varying data dimensionalities arising from different configurations.

We validated Data Aware Differentiable \gls{nas} through experiments on two distinct tasks: the benchmark \gls{gsc} v0.02 dataset and a custom name detection application designed to test flexibility. 
The first experiment allowed comparison with established methods, while the second demonstrated the ease of adapting the framework to new tasks.
Across both experiments, Data Aware Differentiable \gls{nas} successfully identified TinyML systems exhibiting high performance and efficiency (low parameter counts).
These co-optimized systems generally achieved superior accuracy compared to baseline systems built using standard TinyML model architectures evaluated with the same discovered data configurations.
This work demonstrates the potential of holistically optimizing data and architecture choices to push the boundaries of efficient TinyML system design.
Future work could explore gradient descent directly on continuous data options, a new neural architecture search space tailored for TinyML, and additional hyperparameter optimization.

\bibliographystyle{IEEEbib}
\bibliography{bibliography}

\end{document}